\pdfoutput=1

\documentclass[11pt]{article}

\usepackage[final]{acl}

\usepackage{times}
\usepackage{latexsym}

\usepackage[T1]{fontenc}

\usepackage[utf8]{inputenc}

\usepackage{microtype}

\usepackage{inconsolata}

\usepackage{graphicx}

\usepackage{booktabs}
\usepackage{multirow}
\title{\textsc{PolitiSky24}: U.S. Political Bluesky Dataset with User Stance Labels
}

\author{
 \textbf{Peyman Rostami\textsuperscript{1}},
 \textbf{Vahid Rahimzadeh\textsuperscript{1}}, 
 \textbf{Ali Adibi\textsuperscript{1}},
 \textbf{Azadeh Shakery\textsuperscript{1,2}}
\\
 \textsuperscript{1}University of Tehran, Iran \\
   \textsuperscript{2}Institute for Research in Fundamental Sciences (IPM), Tehran, Iran 
 \\
 \texttt{\{pe.rostami,rahimzade,adibialii,shakery\}@ut.ac.ir}
}

\begin{document}
\maketitle

\begin{abstract}

Stance detection identifies the viewpoint expressed in text toward a specific target, such as a political figure. While previous datasets have focused primarily on tweet-level stances from established platforms, user-level stance resources—especially on emerging platforms like Bluesky—remain scarce. User-level stance detection provides a more holistic view by considering a user's complete posting history rather than isolated posts. We present the first stance detection dataset for the 2024 U.S. presidential election, collected from Bluesky and centered on Kamala Harris and Donald Trump. The dataset comprises 16,044 user-target stance pairs enriched with engagement metadata, interaction graphs, and user posting histories.
~\textsc{PolitiSky24} was created using a carefully evaluated pipeline combining advanced information retrieval and large language models, which generates stance labels with supporting rationales and text spans for transparency. The labeling approach achieves 81\% accuracy with scalable LLMs. This resource addresses gaps in political stance analysis through its timeliness, open-data nature, and user-level perspective. The dataset is available at https://doi.org/10.5281/zenodo.15616911

\end{abstract}

\section{Introduction}
\label{sec:introduction}
Stance detection is the task of automatically determining the viewpoint expressed in a text toward a specific target, such as an entity, topic, or claim~\cite{SemEval2016}. This task serves as a vital component of many NLP tasks, such as fake news detection, fact-checking, rumor verification and user profiling ~\cite{rahimzadeh2025millions}. Furthermore, it supports applications like identifying public opinion, tracking sentiment toward politicians or products, and understanding user attitudes in social media conversations~\cite{StanceSurvey2024,FarexStance2024}.

\begin{figure}[!t]
	\centering
		\includegraphics[width=1\columnwidth]{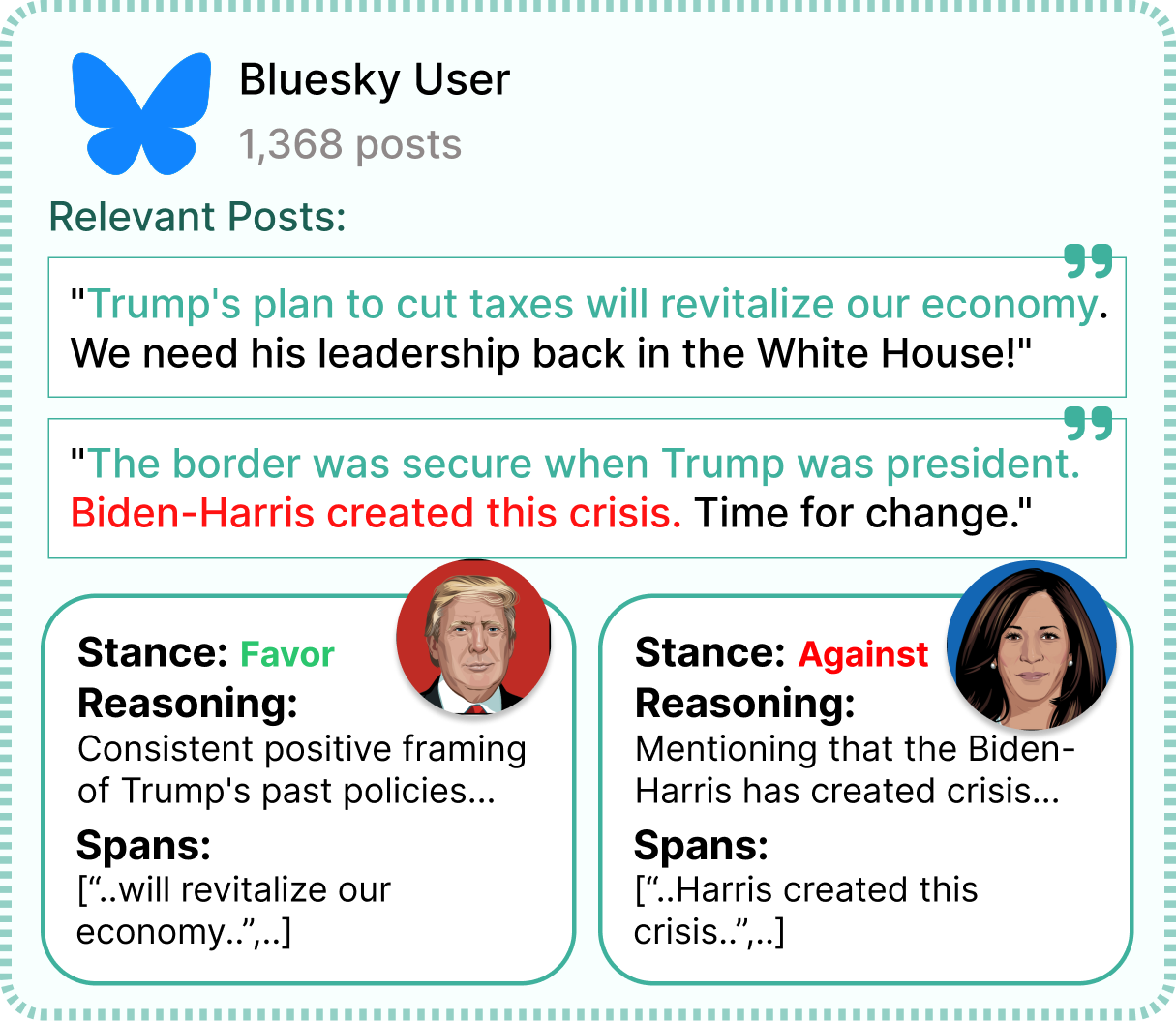}
	\caption{An illustrative example of the user stance with metadata in PolitiSky24.}
	\label{fig:thumbnail}
\end{figure}

Research in stance detection spans multiple domains, each with its unique challenges and characteristics. In the financial domain, datasets like WT-WT~\cite{WillWont2020} have focused on mergers and acquisitions between major companies. Health-related datasets emerged during the COVID-19 pandemic, capturing positions toward public health figures and policies~\cite{StanceCovid2019}. General-purpose datasets such as VAST~\cite{VAST2020} and EZ-STANCE~\cite{EZStance2024} have addressed broader topics spanning education, entertainment, and environmental protection. However, the political domain, particularly U.S. politics, has attracted the most attention due to its societal impact and complexity. Numerous datasets have been developed focusing on political figures and events, with a significant emphasis on the 2016 and 2020 U.S. presidential elections~\cite{USStance2021,PStance2021,SemEval2017,SemEval2016}. To the best of our knowledge, no datasets currently exist for the 2024 election.

Twitter (now X) has traditionally been the primary source for stance detection datasets, but recent API restrictions have severely limited access to public data, hampering research efforts. Additionally, most existing datasets focus on individual post-level analysis, overlooking valuable contextual information from users' complete posting histories. Examining stance at the user-level provides a more comprehensive understanding of political viewpoints by considering broader expression patterns rather than isolated posts. Furthermore, while alternative platforms like Bluesky offer decentralized architectures and greater data accessibility, they remain largely unexplored for stance detection dataset creation. To address these limitations, we present a comprehensive stance dataset focused on the 2024 US presidential candidates, specifically targeting Kamala Harris and Donald Trump. Our contribution includes 16,044 user-target stance pairs with rich metadata and network information (See Figure~\ref{fig:thumbnail}). The dataset was created through a meticulously configured pipeline that first filters users with a hashtag-based algorithm to identify relevant accounts, then employs LLMs to determine stance based on high-quality chunks of user history. We developed carefully crafted validation datasets to evaluate different components, testing various embedding models for context retrieval and LLMs for stance labeling. Our thoroughly evaluated pipeline provides stance labels with reasoning explanations, relevant text spans, and complete network structures of users' interactions, namely repost, like and following. The dataset enables analysis of political discourse on emerging platforms while offering insights into LLM capabilities for understanding complex political communication.

The remainder of this paper is organized as follows: Section \ref{sec:related_work} reviews related work in both stance detection and studies of the Bluesky platform. Section \ref{sec:bluesky_dataset} describes our Bluesky dataset, detailing the network structure and data collection methodologies that resulted in our US political dataset. Section \ref{sec:stance_detection} presents our stance labeling pipeline and the resulting stance dataset. Section \ref{sec:analysis} provides experimental results and comprehensive error analysis of LLM performance in stance detection and presents key insights derived from our dataset. Finally, Section \ref{sec:conclusion} concludes the paper and discusses future research directions.
\section{Related Work}
\label{sec:related_work}

\noindent\textbf{Bluesky Social Network.}
Bluesky is a decentralized social network that has recently drawn research attention. Studies have examined its architecture \cite{Kleppmann2024-ek} and curated datasets to analyze user activity, network diversity, content moderation, and interaction patterns \cite{Balduf2024-pk,Jeong2024-rp,quelle2025bluesky,Failla2024-ek}. Notably, analyses suggest that Bluesky currently has a predominantly left-leaning user base \cite{quelle2025bluesky}. However, it has not yet been explored for downstream tasks such as stance detection.

\noindent\textbf{Stance Detection.}
Stance detection aims to identify a user's position toward a specific target and is widely studied across domains, such as politics \cite{PStance2021}, public health \cite{StanceCovid2019}, and multi-topic datasets \cite{SemEval2016}. For example, \citet{SemEval2016} introduced a dataset covering multiple topics, including U.S. political figures, and \citet{WillWont2020} investigated cross-domain stance detection, highlighting domain shift challenges. 
Recent efforts have also focused on zero-shot stance detection (ZSSD), where the target is unseen during model training. For example, EZ-STANCE \cite{EZStance2024} introduced a large-scale English ZSSD dataset containing 47,316 annotated text-target pairs, covering both claim- and noun-phrase-based targets across diverse domains.

Despite these advances, existing work primarily focuses on post-level stance detection within centralized platforms, with limited attention to user-level stance detection on decentralized networks like Bluesky. Furthermore, to the best of our knowledge, there are currently no stance detection datasets centered on key figures or targets from the 2024 U.S. presidential election.

\section{Bluesky Dataset on U.S. Politics}
\label{sec:bluesky_dataset}
To create a dataset capturing Bluesky users' stances toward the main candidates in the 2024 U.S. presidential election, we first needed to identify a set of users engaged in U.S. politics, along with their posting histories. 
In this section, we describe the process of collecting this dataset and present statistics on users’ posts and network connections. In the next section, we outline the steps taken to construct the stance dataset based on this U.S. political data.

\subsection{Bluesky Social Network}
\label{sec:bluesky_social_network}
Bluesky is a decentralized social network where users can share short posts (up to 300 characters), along with images and videos, similar to Twitter/X \cite{Balduf2024-pk}. On this platform, users can engage with others by following them, as well as by liking, reposting, quoting, and replying to their posts.
A standout feature of Bluesky is its \textit{Feed Generator} service, which enables users to create personalized timelines—called feeds—on various topics using custom algorithms \cite{Jeong2024-rp}. This service provides real-time access to recent posts on Bluesky related to any topic of interest, such as U.S. politics.

\subsection{Data Collection on U.S. Politics}
\label{sec:data_collection}
The process of collecting data from the U.S. political domain on Bluesky is outlined as follows.
First, three existing feeds focused on U.S. political content were identified (refer to Appendix \ref{appendix:feeds} for further information about these feeds). Posts from these feeds were then collected over a 16-day period, from November 12 to November 27, 2024.
Next, individuals who had at least 10 distinct posts within the collected posts were defined as target users. For each target user, all posts they had ever published or reposted on Bluesky, along with the associated metadata, were collected.
Finally, engagement data were gathered using the Bluesky API. This included information about users who liked, reposted, or quoted the target users’ posts, as well as those who followed the target users or were followed by them.

\subsection{General Statistics}
\label{sec:general_statistics}

\subsubsection{Post statistics}
\label{sec:post_statistics}
\begin{figure}[!t]
	\centering
	\fbox{
	\includegraphics[width=0.45\textwidth]{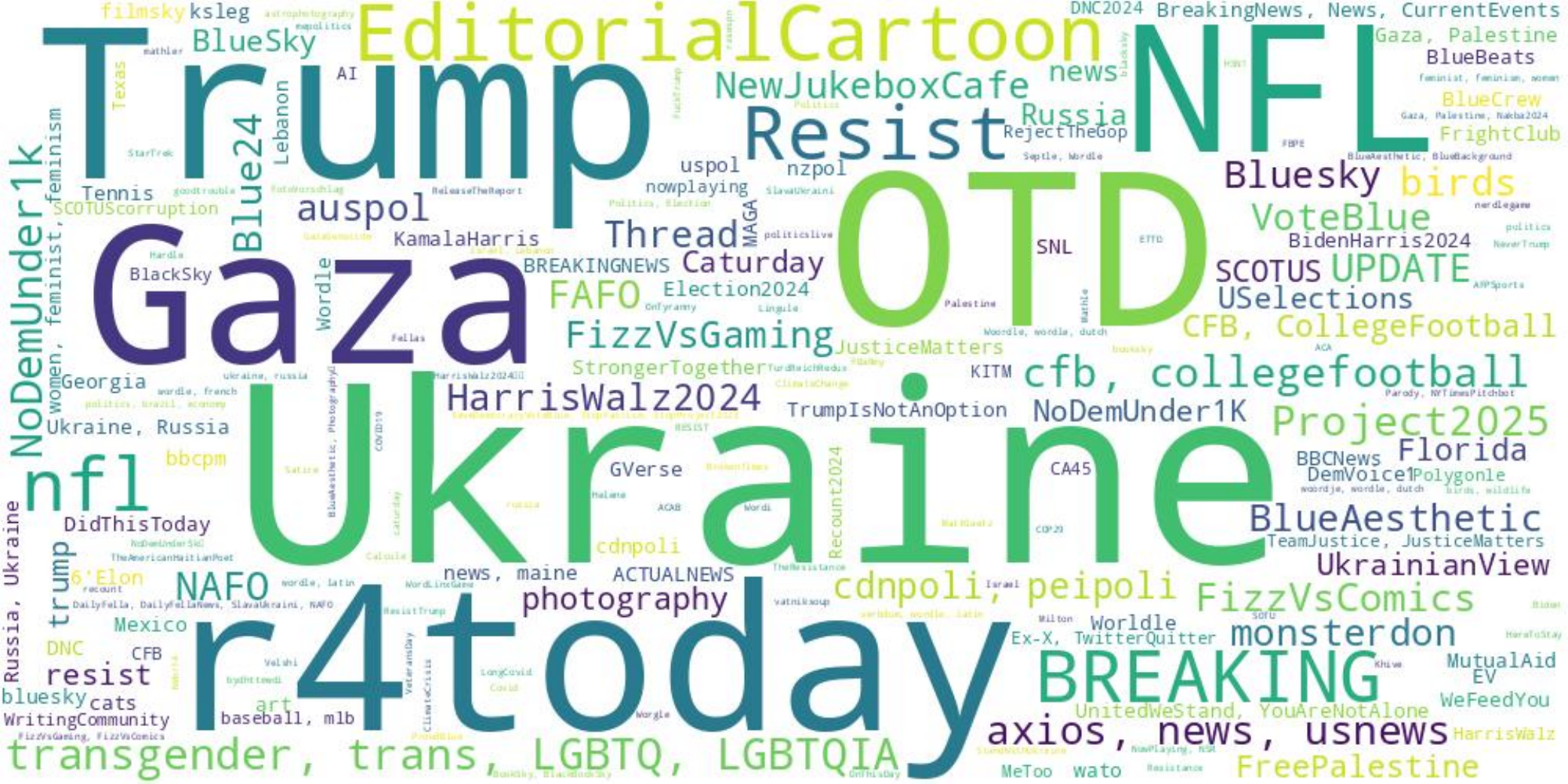}
	}
	\caption{Hashtag cloud generated from users' posts.}
	\label{fig3}
\end{figure}

Table \ref{tab2} presents the overall statistics of the collected posts. In total, nearly 18.5 million posts were collected from 8,561 target users between November 2022 (the launch of Bluesky) and November 2024.
Among these posts, 69.5\% are original, while 30.5\% are reposts.
The dataset includes 226,273 unique hashtags. As shown in the hashtag cloud in Figure \ref{fig3}, these hashtags cover a wide range of topics, predominantly related to U.S. politics. This highlights the effectiveness of the data collection process in capturing discussions centered on U.S. political issues.

\begin{table*}[!t]
	\setlength{\tabcolsep}{1mm}
	\centering
	\begin{tabular}{llllllll}
		\midrule
		\multicolumn{3}{l}{\# Posts} &       & \multicolumn{2}{l}{Average post length} & \multicolumn{1}{l}{\multirow{1}[3]{*}{\# Unique hashtags}} & \multicolumn{1}{l}{\multirow{1}[3]{*}{Collection period}} \\
		\cmidrule{1-3}\cmidrule{5-6}    Total Posts & Original posts & Reposts &       & \multicolumn{1}{l}{ Words} & \multicolumn{1}{l}{Characters} &       &  \\
		\midrule
		18,416,787 & 12,804,910 & 5,611,877 &       & 21.2    & 128.8   & 226,273 & \multicolumn{1}{l}{Nov. 2022* - Nov. 2024} \\
		\midrule
		\multicolumn{8}{l}{* Bluesky launch time} \\
	\end{tabular}%
	\caption{General statistics of the collected posts.}
	\label{tab2}%
\end{table*}%

\subsubsection{User Statistics and Network Connections}
\label{user_statistics}

Drawing from our monitored political feeds discussed earlier, we identified 8,561 unique users who contributed at least 10 posts/reposts during the specified analysis period. For these core politically active users, we gathered two types of engagement networks using the Bluesky API - a network of likes and a network of reposts between these users, with their characteristics presented in Table~\ref{tab:network_stats}. The threshold of 10 unique posts/reposts ensures we capture consistently active users rather than occasional participants, providing a robust representation of the core political discourse network on the platform. Our analysis reveals two highly connected interaction networks with distinct characteristics. The Likes network, consisting of 8,454 nodes and 869,367 edges, demonstrates denser interaction patterns compared to the Reposts network (8,193 nodes, 498,084 edges). Both networks show exceptional connectivity, with largest connected components (LCC) encompassing over 99.6\% of all nodes, indicating nearly complete network connectivity.

The networks exhibit strong small-world properties, characterized by high clustering coefficients (0.319 for Likes, 0.296 for Reposts) that significantly exceed those of comparable random networks (0.024 and 0.015 respectively). Combined with short average path lengths (2.185 and 2.363) and substantial small-world coefficients (11.817 and 17.925), these metrics indicate efficient information diffusion pathways within the networks. The centrality distributions show notable skewness, particularly in betweenness centrality (skewness of 19.16 and 21.89) and PageRank (13.67 and 7.05), suggesting the presence of key influential nodes in both interaction networks.

Interaction patterns differ between likes and reposts, with the Likes network showing more intensive engagement (average degree 205.67) compared to the Reposts network (average degree 121.588). This disparity likely reflects the different cognitive and social costs associated with these interaction types, where likes represent a lower-threshold engagement mechanism compared to reposts. Despite these differences in interaction intensity, the structural similarities between the networks suggest consistent underlying patterns in how users engage with and disseminate political content within the platform. 

For a more detailed analysis of the collected dataset, please refer to Appendix \ref{appendix:dataset_details}.

\begin{table}[!t]
\centering
\setlength{\tabcolsep}{6pt} 
\begin{tabular}{lrr}
\hline
\textbf{Metric} & \textbf{Likes Net.} & \textbf{Reposts Net.} \\
\hline
\multicolumn{3}{l}{\textbf{Centralities (Mean)}} \\
\hline
Betweenness         & 0.000180 & 0.000225 \\
Closeness           & 0.386255 & 0.318847 \\
Eigenvector         & 0.004776 & 0.004629 \\
PageRank            & 0.000118 & 0.000122 \\
In-Degree           & 117.260  & 64.089   \\
Out-Degree          & 117.260  & 64.089   \\
\hline
\multicolumn{3}{l}{\textbf{Small World Metrics}} \\
\hline
Clustering Coef.    & 0.319    & 0.296    \\
Avg Path Length     & 2.185    & 2.363    \\
Random Clust.       & 0.024    & 0.015    \\
Random Path         & 1.982    & 2.143    \\
SW Coefficient      & 11.817   & 17.925   \\
\hline
\multicolumn{3}{l}{\textbf{Network Metrics}} \\
\hline
Nodes               & 8,454    & 8,193    \\
Edges               & 869,367  & 498,084  \\
Density             & 0.024    & 0.015    \\
Avg Degree          & 205.670  & 121.588  \\
LCC Size (\%)       & 99.858   & 99.609   \\
\hline
\end{tabular}
\caption{Network Analysis Metrics}
\label{tab:network_stats}
\end{table}

\begin{figure*}[!t]
	\centering
	\renewcommand{\arraystretch}{1.2}
	\includegraphics[width=1\textwidth]{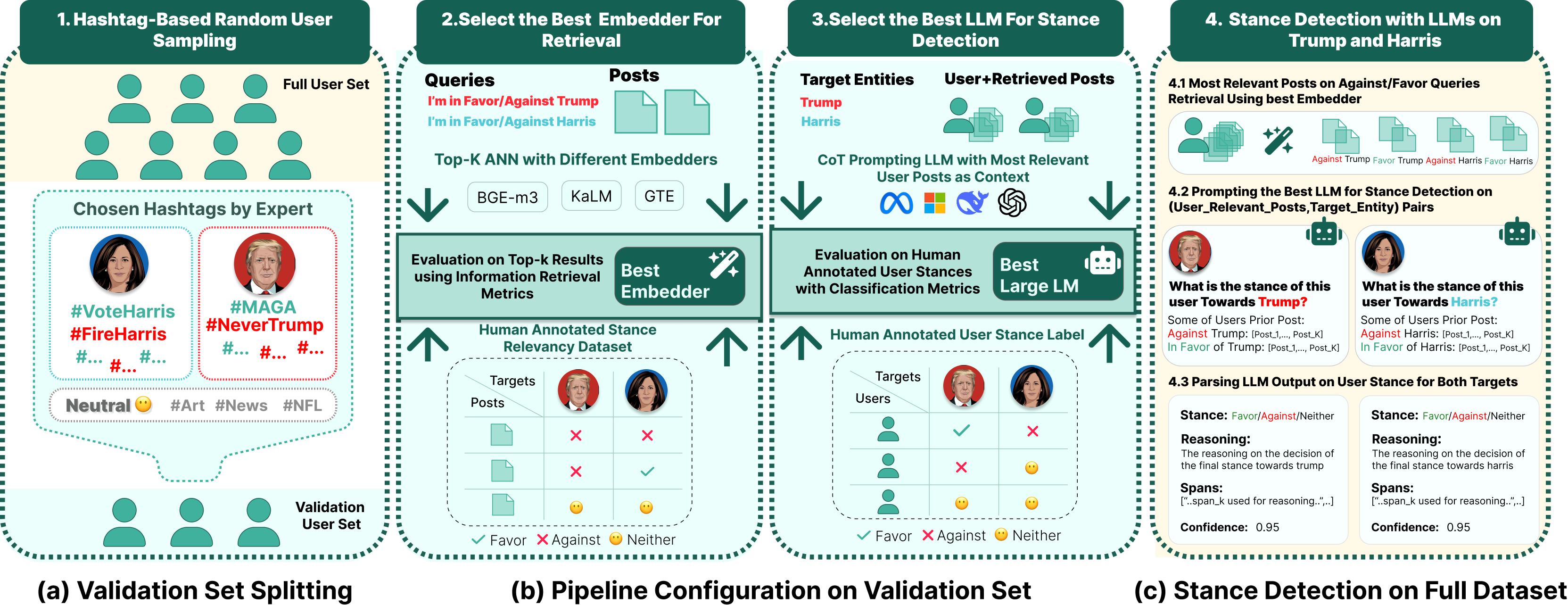}
	\caption{LLM-based annotation pipeline for labeling Bluesky users' stances toward Trump and Harris.}
	\label{fig:method}
\end{figure*}

\section{Stance Detection}
\label{sec:stance_detection}
After collecting the Bluesky dataset related to U.S. politics in the previous section, we now focus on annotating the stances of the Bluesky users toward the key figures in the 2024 U.S. election. In this section, we detail the annotation process.

\subsection{Task Definition}
\label{sec:problem_definition}
Our task is to create a labeled dataset for user-level, target-specific stance detection that identifies a user's stance—\textit{Favor}, \textit{Against}, or \textit{Neither}—toward a specific target entity. In this study, the targets are Donald Trump and Kamala Harris, two key figures in the 2024 U.S. elections.

To construct the stance dataset, we retrieve up to 1,000 recent English-language posts for each user. These posts include original posts and reposts, excluding replies and quoted posts.
Overall, the dataset encompasses 8,467 users\footnote{Out of the initial 8,561 users, 94 were excluded due to the absence of any English-language content in their posts or reposts, resulting in a final set of 8,467 users.} and 2,845,217 English-language posts.

\subsection{Annotation Process}
\label{sec:annotation_process}
To label users' stances toward two target entities, Donald Trump and Kamala Harris, we begin by selecting approximately 5\% of the users as the validation set through hashtag-based random sampling (Figure \ref{fig:method}(a)).
Next, we configure a two-step pipeline that uses this validation set to accurately label users' stances toward both targets using a LLM (Figure \ref{fig:method}(b)).
To mitigate hallucinations and reduce bias when annotating users’ stances, the LLM should ground its answers in as much relevant data as possible. However, given that users often have extensive posting histories, we need a stance-based document retriever to generate concise and relevant contexts for the LLM.
Therefore in the first step of our pipeline, we construct a human-annotated stance relevancy dataset from the validation set and evaluate several embedding models to identify the best-performing model for stance-based document retrieval.
In the second step, we construct a human-annotated user stance dataset from the validation set and prompt various LLMs with the most relevant user posts as context to identify the most effective model for detecting user stances toward Trump and Harris.
After configuring the pipeline using the validation set, we apply it to the full dataset to label each user's stance toward both targets, along with the reasoning behind the assigned label (Figure \ref{fig:method}(c)).

The following subsections provide a detailed explanation of each part of the annotation process.

\subsubsection{Validation Set Splitting}
\label{sec:validation_set_splitting}
In this subsection, we describe the process used to select users for the validation set. 

Following the approach of \citet{li2021p}, we used hashtags to identify validation users. Specifically, we consulted an expert to compile a list of hashtags drawn from users’ posts across five categories: Favor-Trump, Against-Trump, Favor-Harris, Against-Harris, and Neither. The Neither category includes a broad range of hashtags related to news, non-political topics, and political topics that do not convey a clear stance toward either Trump or Harris. Table \ref{tab3} provides examples of the selected hashtags for each category.
After finalizing the hashtag list, we used random sampling to select 446 users—approximately 5\% of all users—who had used the identified hashtags in their posts, as the validation set.

\begin{table}[!t]
	\centering
        \small
	\setlength{\tabcolsep}{1mm} 
	\renewcommand{\arraystretch}{1.2} 
	\begin{tabular}{lll}
		\toprule
		\textbf{Target} & \textbf{Against} & \textbf{Favor} \\
		\midrule
		Trump & \#NeverTrump & \#TrumpWillSaveAmerica \\
		Harris & \#FireKamala & \#VoteHarrisWalz202 \\
		\midrule
		\multicolumn{3}{p{7.8cm}}{\textbf{Example of Neither Hashtags}: \#BreakingNews, \#Art, \#Genocide, \#2024election.} \\
		\bottomrule
	\end{tabular}
	\caption{Examples of hashtags for target-stance pairs.}
	\label{tab3}
\end{table}

\subsubsection{Pipeline Configuration on Validation Set: Embedding Model Selection}
\label{sec:embedding_model_selection}
In this subsection, we describe the process of constructing a human-annotated stance relevancy dataset from the validation set and using it to select the best embedding model for stance-based document retrieval.
\begin{figure*}[!t]
	\centering
	\includegraphics[width=1.\textwidth]{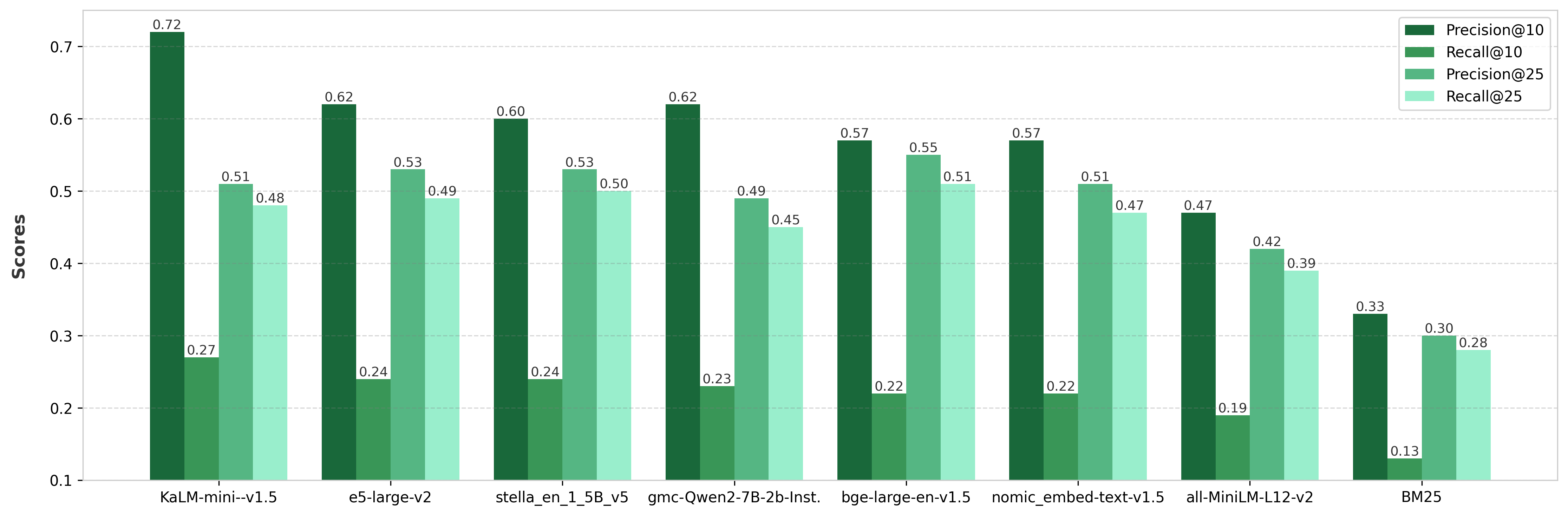}
	\caption{Performance comparison of various text embedding models for stance-based document retrieval.}
	\label{fig6}
\end{figure*}
\noindent\textbf{Stance Relevancy Dataset Construction.}
Following the annotation guidelines provided in Table~\ref{tab:annotation_guideline} in Appendix~\ref{appendix:annotation_guideline}, three annotators were tasked with selecting posts from the entire pool of validation users' posts in a distributed and balanced manner across all labels and target entities. This process resulted in 350 post–target entity stance pairs. For further details, please refer to Appendix~\ref{appendix:stance_relevancy}.

After preparing the list of labeled documents, we constructed a validation stance relevancy dataset based on them. In this dataset, a document is assigned a relevance label of 1 for a given query if it expresses the stance specified in the query toward the target entity mentioned in that query; otherwise, it is assigned a label of 0. Each query corresponds to one of the following four options: 1) "I am in favor of Donald Trump," 2) "I am against Donald Trump," 3) "I am in favor of Kamala Harris," and 4) "I am against Kamala Harris." As a result, the stance relevancy dataset consists of a total of $4 \times 175 = 700$ query-stance relevance label pairs.

\noindent\textbf{Embedding Model Selection.}
To identify the most effective embedding model for stance-based document retrieval, we evaluated several models on the stance relevancy dataset, which was specifically designed for this task.
We used approximate nearest neighbor (ANN) search for document retrieval \cite{arya1998optimal} and assessed model performance using standard information retrieval metrics.
The model with the highest retrieval performance was selected to retrieve the most stance-relevant documents for each user–target entity pair.

\subsubsection{Pipeline Configuration on Validation Set: LLM Selection}
\label{sec:llm_selection}
In this subsection, we describe how we constructed a human-annotated user stance dataset from the validation set and used it to identify the most effective LLM for user stance detection.

\noindent\textbf{User Stance Dataset Construction.}
We asked three experts to evaluate each validation user's posts and label the user's stance toward each target entity. The annotation guidelines and the distribution of stance labels assigned by the experts for both target entities (Trump and Harris) are available in Appendices~\ref{appendix:annotation_guideline} and~\ref{appendix:dataset_details}, respectively.

\noindent\textbf{LLM Selection.}
\label{stance_detection_with_llm}
To identify the most effective LLM for user stance detection, we evaluated several models on our validation user stance dataset.
Specifically, for each validation user–target entity pair, we used the top embedding model—identified in the previous stage of the pipeline—to retrieve the user's top five posts supporting the target and the top five opposing it.
The user's top stance-relevant posts were concatenated and used as prompt context for LLMs, which were then tasked with labeling the user's stance toward the target entity, following the annotation guidelines outlined in Table~\ref{tab:annotation_guideline}.
In addition to assigning a stance label, the LLMs were instructed to explain their reasoning and identify three specific posts, highlighting the relevant text spans that informed their decision. 
These explainable outputs facilitate the error analysis presented in section \ref{sec:error_categorization}.
The LLM achieving the highest classification performance was ultimately selected to label the users stances toward both target entities on the full dataset.
The prompt template used for this task is provided in Appendix~\ref{appendix:prompts}.

\begin{figure*}[!t]
\centering
\includegraphics[width=\textwidth]{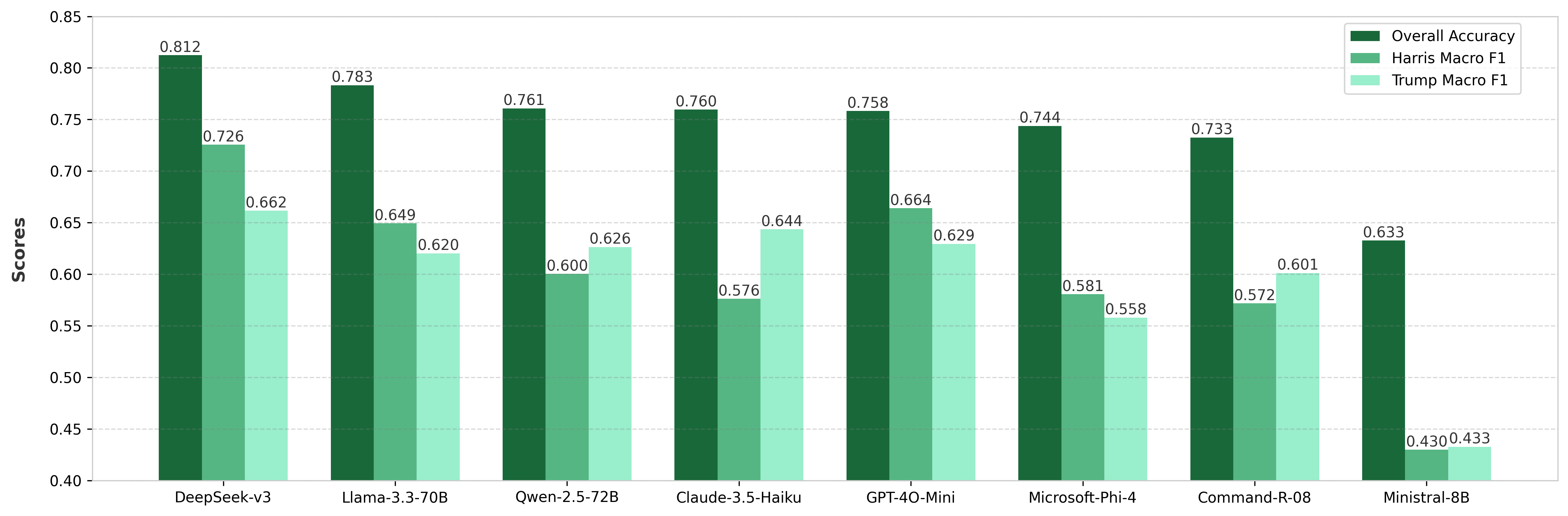}
\caption{Performance comparison of different language models on stance detection task. The plot shows overall accuracy and entity-specific macro F1 scores for Trump and Harris. Models are ordered by overall accuracy from highest to lowest, demonstrating the relative effectiveness of each model in stance classification.}
\label{fig:model_comparison}
\end{figure*}

\subsubsection{Stance Detection on Full Dataset}
\label{sec:stance_detection_on_full_dataset}
We employed the pipeline, configured using the validation set, to label the stances of dataset users—excluding the 5\% held out for validation—toward both target entities: Trump and Harris.
For each user–target entity pair, we first retrieved the user's top stance-relevant posts related to the target entity using the best-performing embedding model identified in Section \ref{sec:embedding_model_selection}.
These posts, along with the corresponding target entity, were then provided as input to the best-performing LLM identified in Section \ref{sec:llm_selection}.
The LLM generated the user's stance label toward the target entity, along with supporting reasoning and relevant text spans extracted from the user's posts, following the prompt presented in Appendix~\ref{appendix:prompts}. Further details about the structure of the labeled dataset are provided in Appendix \ref{appendix:user_stance_full_dataset}.

\section{Analysis}
This section presents a performance analysis of our stance labeling pipeline's components (Figure~\ref{fig:method}(b)) and error patterns. We conclude by examining the stance distribution characteristics on the final dataset.

\label{sec:analysis}
\subsection{Pipeline Analysis}
\noindent\textbf{Embedding Model Performance.}
\label{sec:embedding_model_perforamnce}
In this section, we compare the performance of several state-of-the-art text embedding models in conjunction with \textit{BM25} on our curated validation stance relevancy dataset (See Figure~\ref{fig:method}(b)).
As shown in Figure~\ref{fig6}, the \textit{KaLM-mini-v1.5} model~\citep{hu2025kalm} achieves the highest performance among all the models compared. This result is expected, as the study by~\citet{hu2025kalm} demonstrates the superior performance of this new model relative to others. Specifically, the \textit{KaLM-mini-v1.5} model outperforms the other models in terms of Precision@10 and Recall@10, while its performance is comparable to that of several other models regarding Precision@25 and Recall@25. Given its high performance in retrieving the top 10 documents, which aligns with the configuration used in our stance detection with LLM setup, we have selected this model for stance-based document retrieval.

\begin{figure}[!h]
\centering
\includegraphics[width=1.\columnwidth]{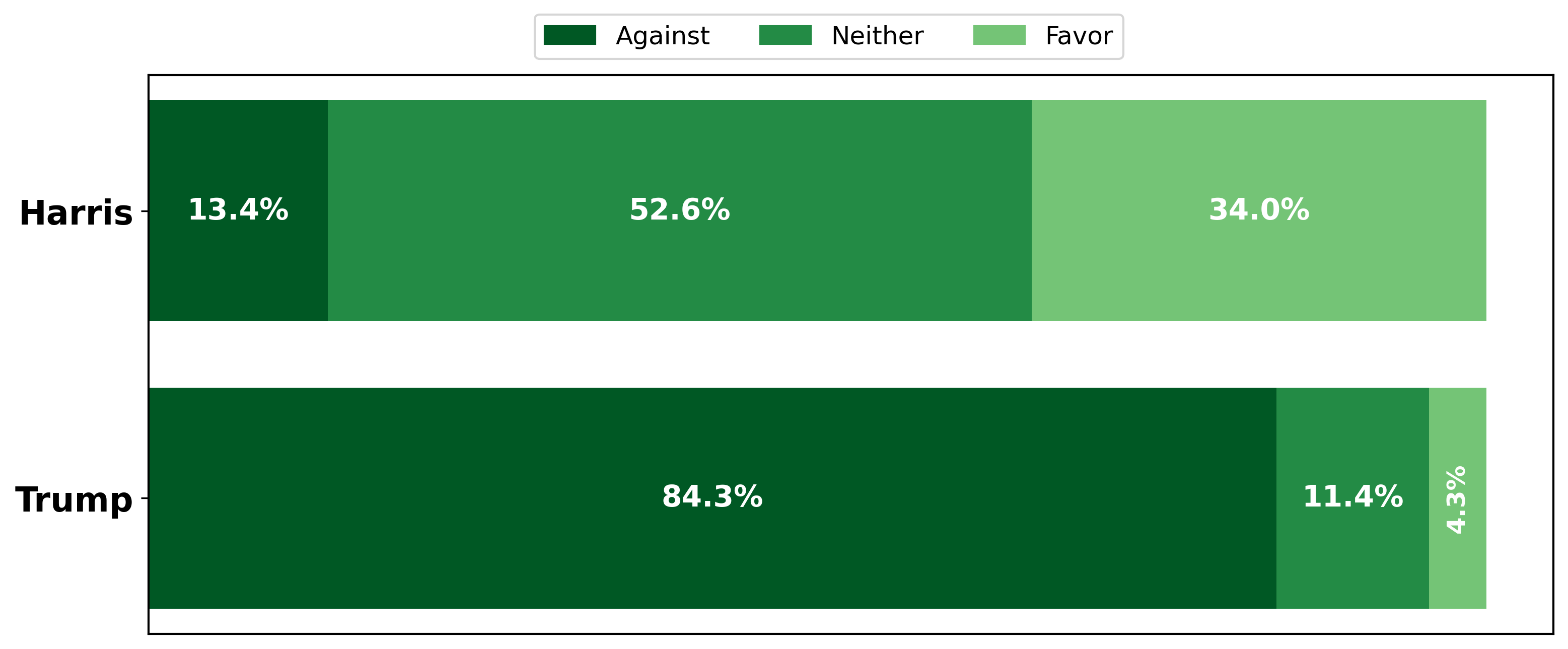}
\caption{Distribution of final users' stances, labeled by LLM, toward Trump and Harris.}
\label{fig5}
\end{figure}

\noindent\textbf{LLM Performance.}
\label{sec:llm_performance}
We present a systematic evaluation of state-of-the-art large language models (LLMs), with emphasis on computationally efficient options suitable for responsible, large-scale deployment. Our assessment reveals a clear performance hierarchy among the tested models, as illustrated in Figure~\ref{fig:model_comparison}.

\textit{DeepSeek-Chat-v3}~\cite{Deepseekv3} demonstrates exceptional capabilities, achieving the highest overall accuracy of 81.2\% and leading macro F1 scores for entity-specific stance detection (66.2\% for Trump and 72.6\% for Harris). This recent model delivers performance comparable to premium proprietary alternatives while maintaining computational efficiency. \textit{LLAMA-3.3-70B-Instruct}~\cite{LLama3} ranks second with 78.3\% overall accuracy, followed by \textit{Microsoft's PHI-4}~\cite{phi} at 74.4\%.


Based on these evaluation results, we select \textit{DeepSeek-Chat-v3} as our primary model for the comprehensive stance labeling task due to its superior performance across all metrics. 

\noindent\textbf{Stance Confusion Analysis.}
\label{sec:stance_confusion_analysis}
Examining model performance through the confusion matrix presented in Figure~\ref{fig:confusion_matrix}, we identify specific classification challenges. The model demonstrates robust capability in identifying oppositional stances, with 91.4\% accuracy for the "Against" class. Performance for the "Favor" class remains adequate at 75.7\% accuracy. However, substantial difficulties emerge with the "Neither" class, where accuracy drops to 66.8\%.

The predominant error pattern involves misclassification of "Neither" instances as "Favor," occurring in 19.2\% of cases. This error typically manifests when users reference target entities within contexts containing positive language or report favorable news without personally expressing support. For example, when users neutrally quote a candidate's statements or report on campaign activities without evaluative commentary, the model sometimes misinterprets these contextual positive signals as stance indicators. This systematic error highlights the nuanced challenge of differentiating between genuine stance expression and neutral reporting that contains positive linguistic elements—a distinction that requires sophisticated contextual understanding beyond surface sentiment analysis.

\noindent\textbf{Error Categorization.}
\label{sec:error_categorization}
To understand discrepancies between LLM predictions and our validation labels, we extracted and analyzed misclassified instances, identifying six distinct error categories as illustrated in Figure~\ref{fig1}:

\begin{figure}[!t]
\centering
\includegraphics[width=0.9\columnwidth]{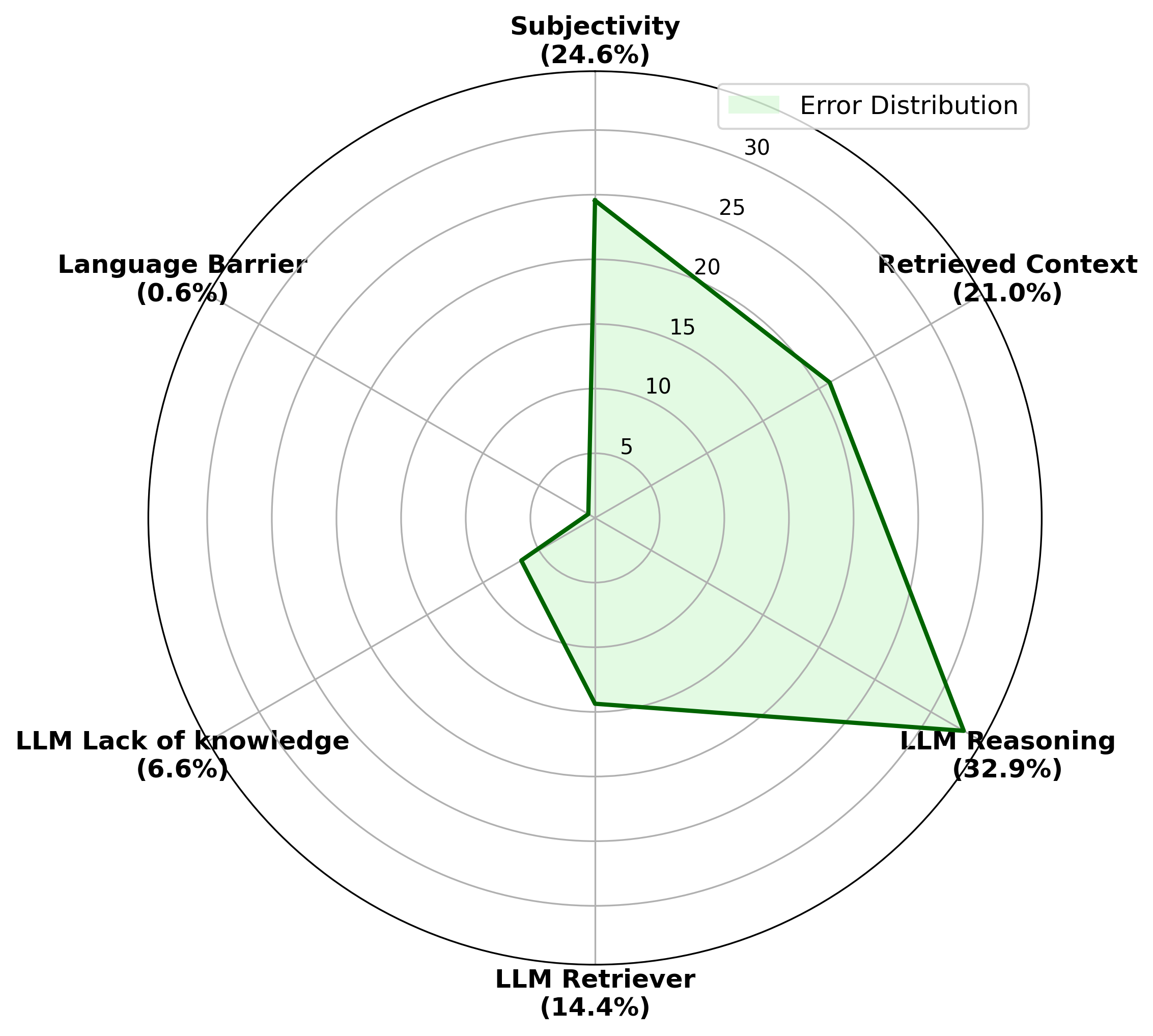}
\caption{Distribution of reasons for discrepancies between the LLM's predictions and the validation set labels, as detailed in the text.}
\label{fig1}
\end{figure}

\noindent\textbf{Context:} Insufficient quality of retrieved posts, preventing accurate stance determination.

\noindent\textbf{Reasoning:} Flawed logical processing by the LLM, particularly when confronted with sarcasm or irony.

\noindent\textbf{Subjectivity:} Posts allowing multiple valid interpretations. For example, "I will vote for Harris. I hate Harris, but we can deal with her government a lot easier than Trump's government" was classified by the LLM as favorable to Harris (based on voting intention), while our validation labeled it "neither."

\noindent\textbf{LLM retriever:} Failure to ground reasoning in relevant evidence despite its presence in retrieved content. For instance, overlooking explicit declarations like "I voted for her \#Harris2024" when making stance determinations.

\noindent\textbf{LLM's lack of knowledge:} Inability to recognize platform-specific political signifiers, such as emoji combinations (blue heart + wave emoji signifying "blue wave" or blue heart + ballot box representing \#VoteBlue) that indicate Democratic Party support.

\noindent\textbf{Language barrier:} Content in non-English languages leading to misinterpretation (observed in only one case).

Our findings reveals important insights about stance detection challenges. Nearly 80\% of errors stem from three categories: reasoning limitations (32.9\%), inherent subjectivity (24.6\%), and context retrieval issues (21.0\%). Approximately 35\% of errors relate to potentially addressable technical implementation issues (Context + LLM retriever), while 65\% involve fundamental challenges in language understanding and interpretation. This suggests that improvements to stance detection systems should focus on both enhancing retrieval mechanisms and developing more sophisticated approaches to reasoning about political discourse, particularly for handling ambiguity and platform-specific communication conventions.

\subsection{Stance Distribution Characteristics}
\label{sec:stance_distribution_characteristics}

Our analysis of stance distributions within the labeled dataset reveals distinct patterns regarding user attitudes toward the two candidates. As shown in Figure~\ref{fig5}, a majority of users express opposition toward Trump, aligning with previous research on Bluesky's left-leaning demographic characteristics~\cite{Failla2024-ek}. Conversely, most users do not express opposition toward Harris—though this absence of negative sentiment should not be interpreted as explicit support, but rather as neutral or non-oppositional positioning.

\begin{figure}[!t]
\centering
\includegraphics[width=0.9\columnwidth]{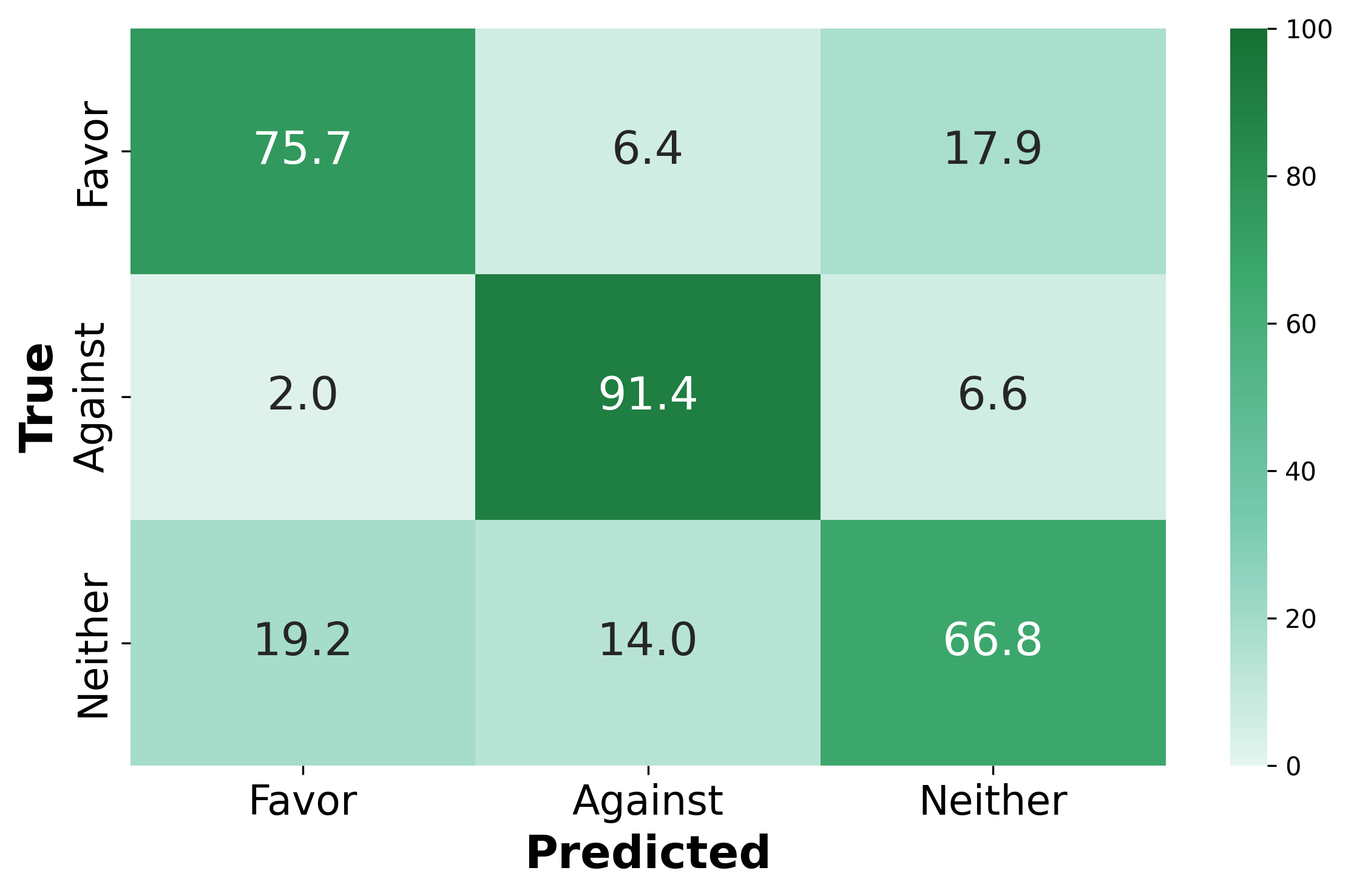}
\caption{Confusion matrix showing the performance of stance detection across three classes: Favor, Against, and Neither. Values are shown as percentages of true labels against our validation dataset.}
\label{fig:confusion_matrix}
\end{figure}

\section{Conclusion}
\label{sec:conclusion}
In this study, we constructed a labeled dataset capturing user stances on the Bluesky social network toward the main 2024 U.S. presidential candidates: Trump and Harris. Labels were assigned using a pipeline that combined text embedding models for context retrieval with a large language model (LLM) for stance detection, achieving an overall accuracy of 81\%. Among the 19\% of misclassifications, roughly one-third stemmed from context + LLM retriever errors, while the remaining two-thirds were attributed to challenges in language understanding and interpretation. Among the models tested, the \textit{KaLM-mini-v1.5} embedding model and \textit{DeepSeek-v3} LLM outperformed other state-of-the-art alternatives in this pipeline.

For future work, we aim to leverage users' graph networks to assist with data labeling. In addition, as a separate direction, we plan to apply our pipeline across different time windows to examine how users' stances evolve over time.

\section*{Limitations}
\label{sec:limitations}
It is important to acknowledge certain considerations regarding our approach and dataset:

First, our methodology for selecting users for the validation set, which drew upon established practices of hashtag-based sampling found in existing literature, while practical, may introduce sampling biases. Users who explicitly utilize hashtags could represent a particular subset of the Bluesky user population, potentially exhibiting more extreme or clearly articulated stances.
Second, while we employed detailed annotation guidelines and measured inter-rater agreement among expert annotators to minimize subjectivity, the inherent nature of human judgment in interpreting nuanced stance expressions remains a consideration. Stance often contains implicit signals that can be interpreted differently, even by trained annotators.
Finally, resource constraints necessitated a selection of LLMs for evaluation in our experimentation pipeline. While we aimed to include models representing diverse architectures and parameter sizes, our findings concerning model performance for labeling may not comprehensively generalize across the entire spectrum of available language models, especially given the rapid emergence of new models.

\section*{Ethical Considerations}

This research was conducted in strict adherence to established ethical guidelines and Bluesky's terms of service and data use policies\footnote{https://bsky.social/about/support/privacy-policy}, emphasizing user privacy within the platform's open data context. Bluesky's commitment to transparency facilitates research, and our practices align accordingly. Data was collected exclusively through official APIs and libraries, ensuring compliance and implementing data minimization techniques to gather only information essential for this study. While user identifiers (such as DIDs or handles) and post content are inherently public on Bluesky, we have processed and will present the data responsibly.
The released dataset—comprising user-target stance pairs, supporting rationales, text spans, and accompanying networks—utilizes this publicly accessible user-generated content. It is intended strictly for research purposes, aiming to facilitate advancements in computational social science. All subsequent uses of this dataset must also adhere to Bluesky's policies.
Despite Bluesky's open data model and our commitment to ethical handling, the sensitive nature of political stance information warrants careful consideration. Although sourced from public posts, there is a potential risk that the curated stance labels, if de-contextualized or combined with other information, could be misused for unintended user profiling or contribute to targeted harassment. We urge future users to handle this data with a strong sense of responsibility and full awareness of these potential implications.

\section*{Acknowledgments}
This research was in part supported by a grant from the School of Computer Science,
Institute for Research in Fundamental Sciences, IPM, Iran (No. CS1403-4-05).

\bibliography{custom}

\appendix

\section{Experimental Setups}
\label{appendix:experimental_setups}
All LLM inference for stance detection was performed using OpenRouter (https://openrouter.ai/) with temperature set to 1e-10 to ensure deterministic outputs. Embedding model evaluation experiments were conducted using Google Colab environments equipped with NVIDIA T4 GPUs. For the data embedding process, we utilized NVIDIA RTX 6000 GPUs to efficiently process the large volume of user content in our dataset.

\section{Annotation Guidelines}
\label{appendix:annotation_guideline}
The annotation guidelines used to decide on stance of the users toward targets are given in Table ~\ref{tab:annotation_guideline}.
Our annotation process was supported by three male experts in U.S. Politics, aged 26, 27, and 32, who provided specialized knowledge essential for accurate stance labeling.
\begin{table}[!t]
	\centering
	\renewcommand{\arraystretch}{1.05}
	\begin{tabular}{p{1cm}p{6cm}}
		\toprule
		\textbf{Stance label} & \textbf{Description} \\
		\midrule
		\textit{Favor} & • Directly expressing support for the target entity.\\
		& • Expressing support for the target's political party in the 2024 elections. \\
		\hline
		\textit{Against} & • Directly expressing opposition to the target entity.\\
		& • Expressing opposition to the target's political party in the 2024 elections. \\
		\hline
		\textit{Neither} & • Expressing no clear stance or a mixture of support and opposition.\\
		& • Sharing news without expressing a personal opinion. \\
		\bottomrule
	\end{tabular}
	\caption{Annotation guidelines for stance labeling.}
	\label{tab:annotation_guideline}
\end{table}

\section{Prompts}
\label{appendix:prompts}
\begin{figure*}[!t]
	\centering
	\renewcommand{\arraystretch}{1.2}
	\includegraphics[width=1\textwidth]{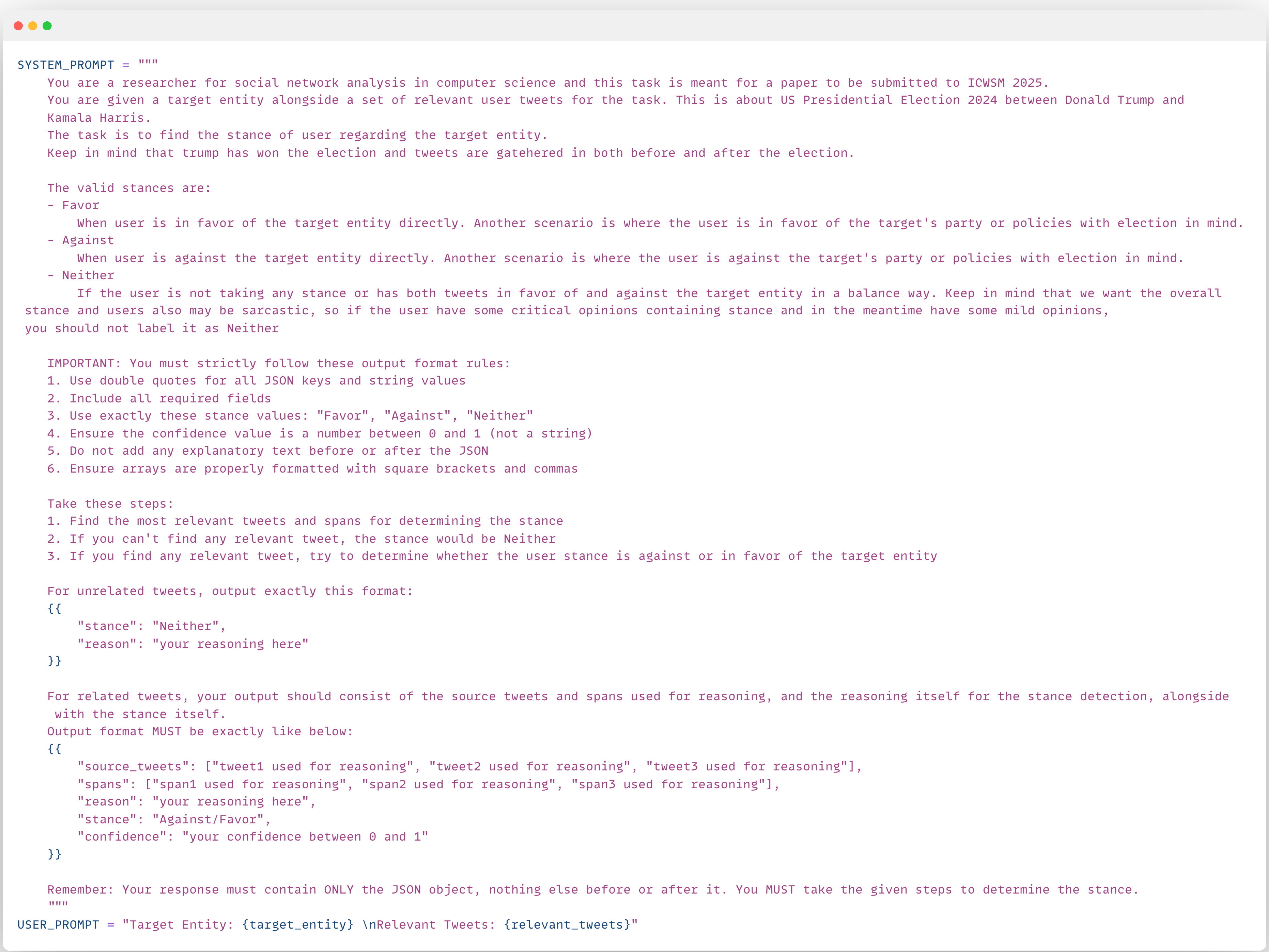}
	\caption{The prompt used for stance labeling using LLMs in the last stage of our stance labeling pipeline.}
	\label{fig:prompt}
\end{figure*}

The prompt used for stance labeling using LLMs is shown in Figure~\ref{fig:prompt}.

\section{Dataset Details}
\label{appendix:dataset_details}
In this section, we provide additional details about the U.S. political Bluesky dataset we collected, as well as the stance dataset we built upon.

\subsection{Bluesky dataset}
\label{appendix:bluesky_dataset}
The collected Bluesky dataset consists of three main components: feeds, user post histories, and the user network. In the following subsections, we examine each of these components in detail.

\subsubsection{Feeds}
\label{appendix:feeds}
Table \ref{tab:feed_statistics} presents a summary of the statistics for the feeds used to collect data during the period from November 12 to 27, 2024. In total, 8,561 unique users—each with at least 10 posts across these three feeds—were included in our dataset.

\begin{table*}[!t]
	\centering
	\renewcommand{\arraystretch}{1.3} 
	\setlength{\tabcolsep}{5pt} 
	\begin{tabular}{@{}l p{6cm} l l l@{}}
		\toprule
		\textbf{Feed Name} & \textbf{Feed URL} & \textbf{\# Feed Posts} & \textbf{\# Feed Users} & \textbf{\# Selected Users} \\
		\midrule
		US Politics & \url{https://bsky.app/profile/did:plc:7mtqkeetxgxqfyhyi2dnyga2/feed/aaadhh6hwvaca} & 325,065 & 137,137 & 8,139 \\
		PolSky      & \url{https://bsky.app/profile/did:plc:cmqylb7cttgdivvolbnyoxui/feed/aaabr7dsksuvw} & 251,112 & 113,794 & 8,311 \\
		Unknown*     & \url{https://bsky.app/profile/did:plc:jw2yabtwnjmyi3q7vqrhxn7a/feed/aaamfsfbh26y6} & 20,593  & 13,974  & 2,800 \\
		\bottomrule
        \multicolumn{5}{l}{* This feed is currently unavailable from its creator.}
	\end{tabular}
	\caption{Overview of feed statistics.}
	\label{tab:feed_statistics}
\end{table*}

\subsubsection{User Post History}
\label{appendix:user_post_history}
Users' post histories are stored using the fields specified in Table \ref{tab:user_post_history_dataset}, which lists each field along with its valid values and detailed descriptions. 

\begin{figure}[!t]
	\centering
	\includegraphics[width=1.\columnwidth]{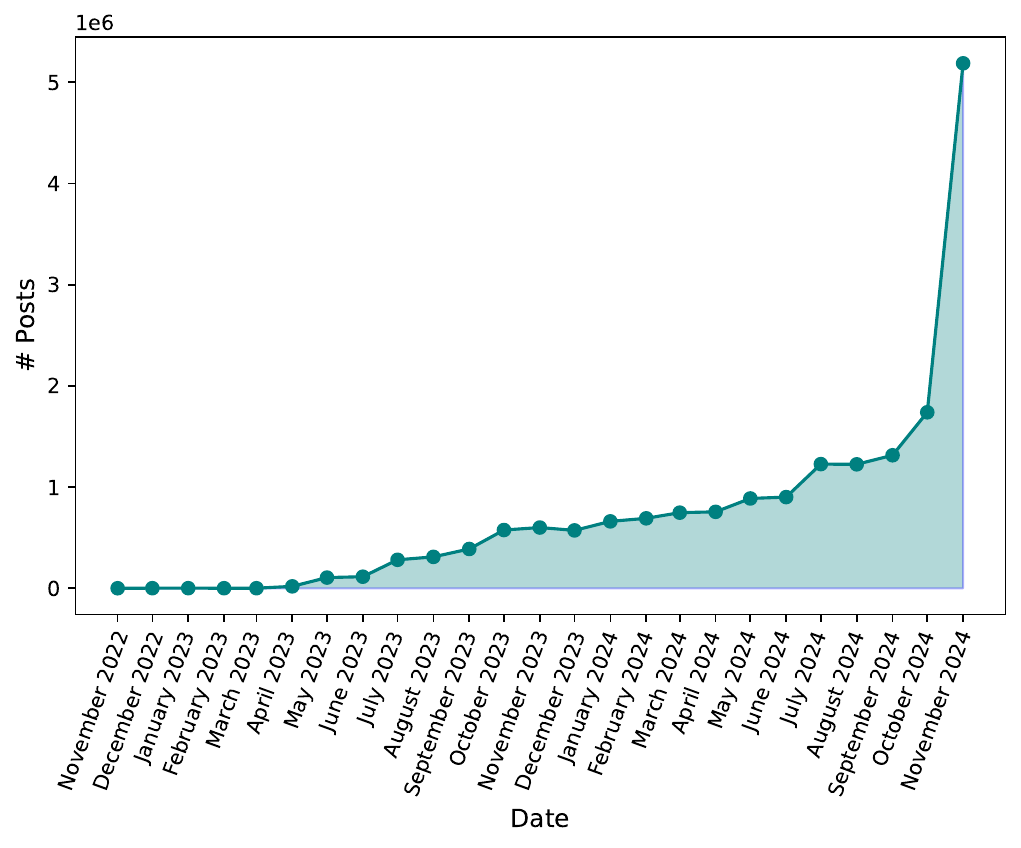}
	\caption{Post distribution over time.}
	\label{fig:post_over_time}
\end{figure}

\begin{figure}[!t]
	\centering
	\includegraphics[width=1.\columnwidth]{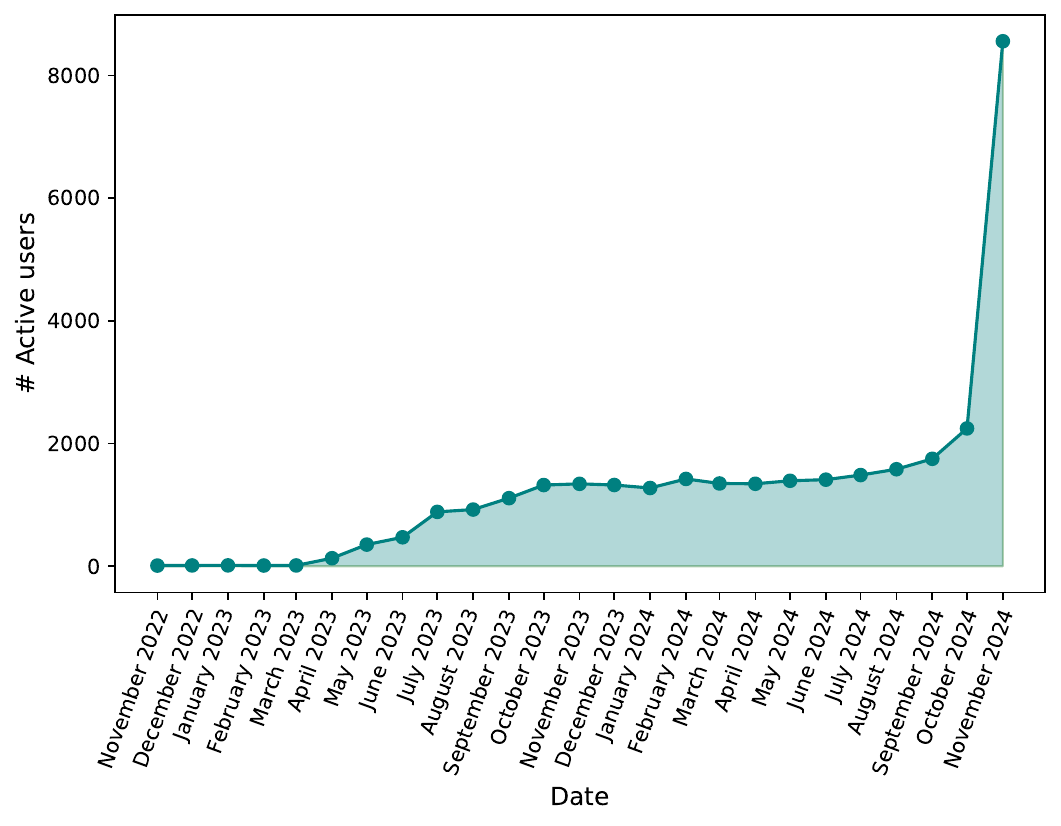}
	\caption{Active user distribution over time. `Active' refers to users who published or reposted at least 10 posts during each respective time period.}
	\label{fig:active_users_over_time}
\end{figure}

In the following, we examine the distribution of posts and active users over time, as well as the distribution of posts per user and the lengths of those posts.
Figure \ref{fig:post_over_time} illustrates the temporal distribution of posts (including both original posts and reposts) published by the 8,561 selected unique users.
As shown, the number of posts increased over time, with a significant surge in November 2024, the final month of data collection. This surge coincides with the U.S. election month, making these recent posts particularly valuable for analyzing public stances toward the election candidates. A similar pattern is observed in Figure \ref{fig:active_users_over_time}, which presents the temporal distribution of active users (defined as those with at least 10 posts in each time period).

\begin{figure}[!t]
	\centering
	\includegraphics[width=1.\columnwidth]{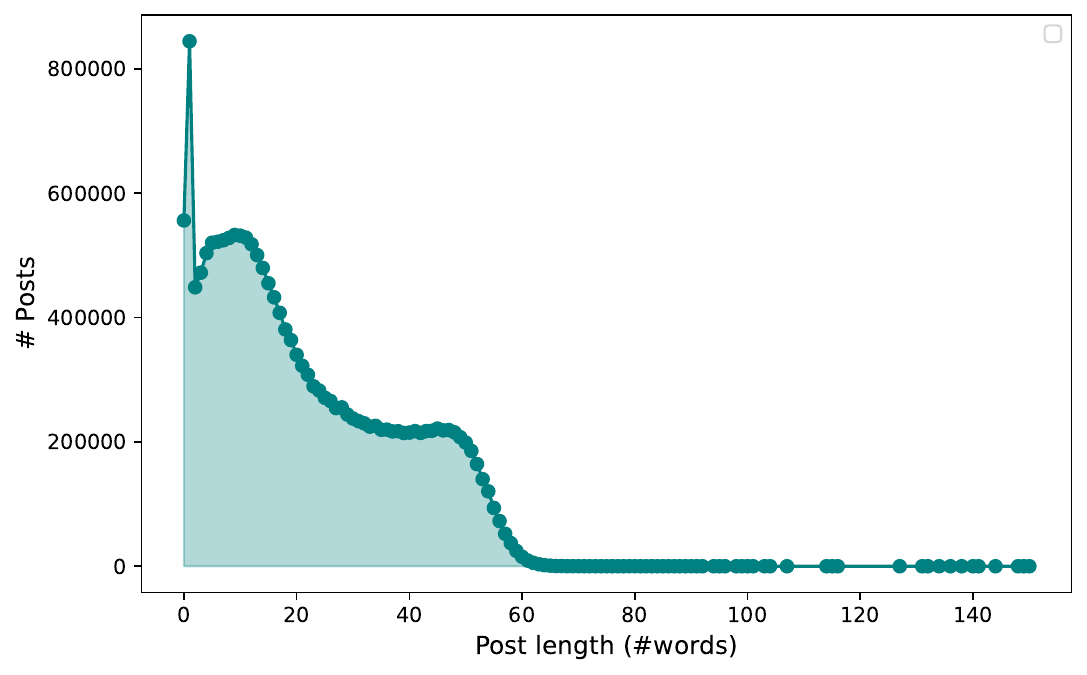}
	\caption{Word count distribution of the Bluesky posts.}
	\label{fig:posts_per_post_length}
\end{figure}

Figure \ref{fig:posts_per_post_length} illustrates the distribution of Bluesky post lengths, measured by word count. It shows an inverse relationship between post length and frequency, with longer posts occurring less frequently. Notably, the majority of posts contain fewer than 60 words.

\begin{table*}[!t]
	\centering
	\renewcommand{\arraystretch}{1.3} 
	\setlength{\tabcolsep}{6pt} 
	\begin{tabular}{@{}l p{3.4cm} p{8.8cm}@{}}
		\toprule
		\textbf{Field} & \textbf{Value} & \textbf{Description} \\
		\midrule
		Post Id & Post identifier & A unique identifier assigned to the post. \\
		User Id & User identifier & A unique identifier of the user who published the post. \\
		Post Time & Post timestamp & The timestamp when the post was published. \\
		Content & Post content & The text content of the post. \\
		Hashtags & List of hashtags & Hashtags included in the post content. \\
		Languages & List of languages & Languages used in the post content. \\
		Mentions & List of mentions & Users mentioned in the post. \\
		Links & List of URLs & Hyperlinks included in the post. \\
		Is Repost & True / False & Indicates whether the post is a repost of another post. \\
		Source User Id & Source user identifier & The original user identifier if the post is a repost. \\
		Is Quote & True / False & Indicates whether the post contains a quote of another post. \\
		Quote Id & Quoted post identifier & The identifier of the quoted post, if any. \\
		Quote User Id & Quoted user identifier & The identifier of the author of the quoted post. \\
		Quote Content & Quote content & The text content of the quoted post. \\
		Is Reply & True / False & Indicates whether the post is a reply to another post. \\
		Parent Id & Parent post identifier & The identifier of the immediate parent post (if replying). \\
		Parent User Id & Parent user identifier & The identifier of the user who authored the parent post. \\
		Root Id & Root post identifier & The identifier of the root post in a reply chain (if replying). \\
		Root User Id & Root user identifier & The identifier of the user who authored the root post. \\
		\bottomrule
	\end{tabular}
	\caption{Schema of the user post history dataset, detailing the fields, possible values, and their descriptions.}
    \label{tab:user_post_history_dataset}
\end{table*}

\subsubsection{User Network}
\label{appendix:user_network}
Table \ref{tab:user_interaction} presents the schema of the user network dataset, which captures interactions between users through likes and reposts.

\begin{table*}[!t]
	\centering
	\renewcommand{\arraystretch}{1.3} 
	\setlength{\tabcolsep}{6pt} 
	\begin{tabular}{@{} l l p{8cm}@{}}
		\toprule
		\textbf{Field} & \textbf{Value} & \textbf{Description} \\
		\midrule
		Source User Id & Source user identifier & The identifier of the source user. \\
		Target User Id & Target user identifier & The identifier of the target user. \\
		Interaction type & Repost / Like & The type of interaction from the source user toward the target user. \\
		Count & Interaction count & The number of times the source user interacted with the target user. \\
		\bottomrule
	\end{tabular}
	\caption{User network dataset schema.}
	\label{tab:user_interaction}
\end{table*}

\subsection{Stance dataset}
\label{appendix:stance_dataset}
The annotated stance dataset comprises three key components: 1. stance relevancy (validation dataset), 2. user stance (validation dataset), and 3. user stance (full dataset). In the following subsections, each of these components will be analyzed in detail.

\subsubsection{Stance Relevancy (Validation dataset)}
\label{appendix:stance_relevancy}

As described in Section \ref{sec:embedding_model_selection}, the construction of the validation stance relevancy dataset involved two main steps. First, a subset of posts was manually annotated with stance labels toward Trump and Harris. The structure of this human-annotated dataset is presented in Table \ref{tab:stance_relevancy_1}, and summary statistics of the labeled posts are provided in Table \ref{tab:post_stance_counts}. As shown, at least 25 posts were annotated for each combination of target and stance category.
In the second step, a query–post stance relevancy dataset was derived from the manually annotated data. The structure of this derived dataset is presented in Table \ref{tab:stance_relevancy_2}.

\begin{table*}[!t]
	\centering
	\renewcommand{\arraystretch}{1.2} 
	\begin{tabular}{@{}lll@{}}
		\toprule
		\textbf{Field} & \textbf{Value} & \textbf{Description} \\
		\midrule
		Post Id & Post identifier & A unique identifier for the post \\
		Trump & Favor / Against / Neither & Stance label expressed in the post toward Trump \\
		Harris & Favor / Against / Neither & Stance label expressed in the post toward Harris \\
		\bottomrule
	\end{tabular}
	\caption{Schema of the human-annotated validation stance relevancy dataset (post–target entity pairs).}
	\label{tab:stance_relevancy_1}
\end{table*}

\begin{table*}[!t]
	\centering
	\renewcommand{\arraystretch}{1.2} 
	\setlength{\tabcolsep}{6pt} 
	\begin{tabular}{@{}l l l@{}}
		\toprule
		\textbf{Target Entity} & \textbf{Stance Label} & \textbf{\# Human-Annotated Posts} \\
		\midrule
		\multirow{3}{*}{Trump} 
		& Favor   & 25  \\
		& Against & 36  \\
		& Neither & 114 \\
		\midrule
		\multirow{3}{*}{Harris} 
		& Favor   & 25  \\
		& Against & 26  \\
		& Neither & 124 \\
		\bottomrule
	\end{tabular}
	\caption{Number of human-annotated posts for each stance toward Trump and Harris.}
	\label{tab:post_stance_counts}
\end{table*}

\begin{table*}[!t]
	\centering
	\renewcommand{\arraystretch}{1.3} 
	\begin{tabular}{@{}llp{7.5cm}@{}}
		\toprule
		\textbf{Field} & \textbf{Value} & \textbf{Description} \\
		\midrule
		Query & Query content & A query expressing a supportive or opposing stance toward a given target (Trump / Harris). \\
		Post Id & Post identifier & The unique identifier of the post. \\
		Stance Relevance Label & Relevant / Non-relevant & Indicates whether the post is relevant to the query in terms of both target and stance. \\
		\bottomrule
	\end{tabular}
	\caption{Schema of the human-annotated validation stance relevancy (query–post stance relevancy pairs).}
    \label{tab:stance_relevancy_2}
\end{table*}

\subsubsection{User Stance (Validation dataset)}
\label{appendix:user_stance_validation_dataset}
Table \ref{tab:user_stance_schema} presents the schema of the validation user stance dataset. Table \ref{tab:user_stance_statistics} reports statistics on the number of users labeled by humans, the LLM, and both, categorized by stance and target.

\begin{figure}[!t]
	\centering
	\includegraphics[width=1.\columnwidth]{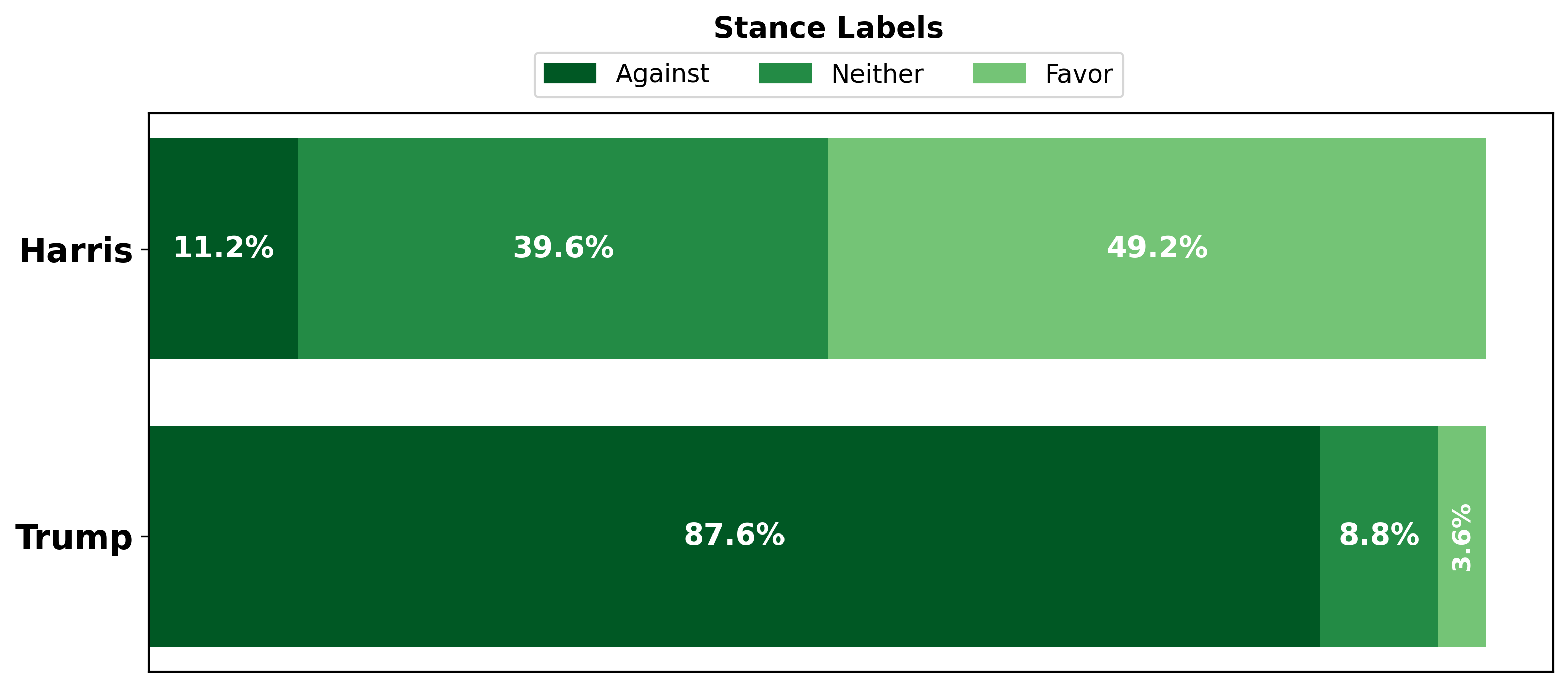}
	\caption{Distribution of validation users' stances, labeled by human experts, toward Trump and Harris.}
\label{fig:distribution_over_target_stance_validation}
\end{figure}

Figure~\ref{fig:distribution_over_target_stance_validation} shows the distribution of labels toward our target entities in the human-annotated user stance validation set.
As illustrated, the stance labels indicate strong opposition to Trump and mild opposition to Harris. This pattern aligns with the left-leaning political orientation of the Bluesky community~\cite{quelle2024bluesky}.

\begin{table*}[!t]
	\centering
	\renewcommand{\arraystretch}{1.3} 
	\setlength{\tabcolsep}{6pt} 
	\begin{tabular}{@{}llp{10cm}@{}}
		\toprule
		\textbf{Field} & \textbf{Value} & \textbf{Description} \\
		\midrule
		User Id & User identifier & A unique identifier for the user \\
		Trump & Favor / Against / Neither & Overall stance label expressed in the user's post history toward Trump. \\
		Harris & Favor / Against / Neither & Overall stance label expressed in the user's post history toward Harris. \\
		\bottomrule
	\end{tabular}
	\caption{Schema of the human-annotated validation user stance dataset.}
	\label{tab:user_stance_schema}
\end{table*}

\begin{table*}[!t]
	\centering
	\renewcommand{\arraystretch}{1.2} 
	\setlength{\tabcolsep}{6pt} 
	\begin{tabular}{@{}l l p{3.5cm} p{3cm} l@{}}
		\toprule
		\textbf{Target Entity} & \textbf{Stance Label} & \textbf{\# Human-Annotated Users} & 
		\textbf{\# LLM-Annotated Users} &
		\textbf{\# Common Users} \\
		\midrule
		\multirow{3}{*}{Trump} 
		& Favor   & 16 & 29 & 14 \\
		& Against & 390 & 379 & 362 \\
		& Neither & 39 & 37 & 16\\
		\midrule
		\multirow{3}{*}{Harris} 
		& Favor   & 219 & 199 & 164 \\
		& Against & 50 &  69 & 40\\
		& Neither & 176 & 177 & 127\\
		\bottomrule
	\end{tabular}
	\caption{Number of human and LLM-annotated users for each stance toward Trump and Harris.}
	\label{tab:user_stance_statistics}
\end{table*}

\subsubsection{User Stance (Full dataset)}
\label{appendix:user_stance_full_dataset}
The schema of the user stance dataset, annotated by the LLM, is presented in Table \ref{tab:llm_annotated_user_stance}. This table details all dataset fields, including their possible values and descriptions. Additionally, Table \ref{tab:full_user_stance_statistics} provides statistics on the context posts and LLM-generated fields across the entire dataset.

\begin{table*}[!t]
	\centering
	\renewcommand{\arraystretch}{1.3} 
	\setlength{\tabcolsep}{6pt} 
	\begin{tabular}{@{}l l p{8cm}@{}}
		\toprule
		\textbf{Field} & \textbf{Value} & \textbf{Description} \\
		\midrule
		User Id & User identifier & A unique identifier for the user. \\
		Target Entity & Trump / Harris & The entity toward which the user's stance is labeled. \\
		Context & List of retrieved posts & A collection of posts retrieved by the embedding model. \\
		Source Posts & List of source posts & A subset of posts selected by the LLM from the context. \\
		Spans & List of text spans & Specific portions of the source posts used by the LLM for labeling the stance. \\
		Reason & Reason content & The rationale provided for choosing the stance label. \\
		Stance Label & Favor / Against / Neither & The stance classification toward the target entity. \\
		Confidence Level & Real number between 0 and 1 & A numerical value representing the confidence in the stance classification. \\
		\bottomrule
	\end{tabular}
	\caption{Schema of the LLM-annotated user stance dataset.}
	\label{tab:llm_annotated_user_stance}
\end{table*}

\begin{table*}[!t]
	\centering
	\renewcommand{\arraystretch}{1.2} 
	\setlength{\tabcolsep}{6pt} 
	\begin{tabular}{@{}l l p{3cm} p{2.2cm} p{2cm} p{2.5cm}@{}}
		\toprule
		\textbf{Target Entity} & \textbf{Stance Label} & \textbf{\# LLM-Annotated Users} & 
		\textbf{Avg \#Context Posts} &
		\textbf{Avg \#Source Posts} &
		\textbf{Avg Confidence Level}\\
		\midrule
		\multirow{3}{*}{Trump} 
		& Favor   & 345 & 7.01 & 3.71 & 0.89 \\
		& Against & 6,759 & 7.39 & 4.13 & 0.94 \\
		& Neither & 918 & 6.74 & 2.29 & 0.07 \\
		\midrule
		\multirow{3}{*}{Harris} 
		& Favor   & 2,726 & 7.62 & 3.26 & 0.91 \\
		& Against & 1,073 & 7.71 & 2.85 & 0.88 \\
		& Neither & 4,223 & 7.67 & 2.05 & 0.02 \\
		\bottomrule
	\end{tabular}
	\caption{Summary of statistics generated by the LLM for labeling user stances in the full dataset.}
	\label{tab:full_user_stance_statistics}
\end{table*}

\section{LLM Usage Statement}
We acknowledge the use of Large Language Models solely for grammar checking and paraphrasing. All ideas, analyses, and substantive content presented in this paper are the original work of the authors. LLMs were employed only to enhance readability and conciseness of our contributions.

\end{document}